\documentclass{article}
\usepackage{spconf,amsmath,graphicx,hyperref}

\usepackage{booktabs}
\usepackage{multirow}
\usepackage{graphicx}
\usepackage{array}
\usepackage{makecell}
\usepackage{xcolor}
\usepackage[numbers]{natbib}

\usepackage{colortbl} % 用于行背景色
\usepackage{pifont}
\usepackage{amsfonts}
\definecolor{creamavg}{HTML}{FFF8E1}
\definecolor{creamavgdark}{HTML}{FBEFC4}

\title{MyoFlow: Anchor-Tied Rectified Flow for HD-sEMG Gesture Recognition Across Sessions and Subjects}
\name{%
  \shortstack{
    Chenhao Wu$^{1}$, Dingjie Peng$^{1}$\sthanks{Corresponding author}, Zhihe Zhang$^{1}$, Satoshi Funabashi$^{1}$, Satoshi Konishi$^{2}$\\ Wuqiang Yang$^{3}$, Hiroshi Onoda$^{1}$, Hironori Washizaki$^{1}$, Jiang Liu$^{1}$
  }%
}
\address{$^{1}$ Waseda University, Japan\quad$^{2}$ KDDI Research Inc., Japan\quad$^{3}$ The University of Manchester, UK}
\begin{document}

\maketitle
\begin{abstract}
High-density surface electromyography (HD-sEMG) gesture recognition supports prosthetic control, assistive robotics, and rehabilitation, but electrode re-donning and physiological variability cause distribution shifts that degrade accuracy across sessions and subjects. Generative HD-sEMG models primarily synthesize signals for augmentation; although diffusion models enhance representation learning, prediction still relies on a separate classifier. To tie learned dynamics to the decision rule, we propose MyoFlow, the first discriminative flow-matching framework for HD-sEMG recognition across sessions and subjects. It recasts classification as anchor-tied transport: a domain-conditioned rectified flow moves encoded windows toward gesture anchors that serve as transport targets and define the nearest-anchor decision geometry, enabling zero-shot recognition without an independent head. On the Hyser dataset, MyoFlow improves mean cross-session and cross-subject accuracy over the strongest diffusion-based baseline by 4.24\% and 6.37\%, respectively, and achieves 91.71\% mean zero-shot accuracy and 97.39\% mean few-shot accuracy across multiple days on the CEMHSEY dataset.
\end{abstract}

\section{Introduction}
\label{sec:intro}
High-density surface electromyography (HD-sEMG) records spatially resolved muscle activity through dense electrode arrays, providing a noninvasive interface for decoding motor intent~\cite{li2021prostheses,yang2025assistive}. Models trained in a single recording domain, however, often generalize poorly to a new session or an unseen user~\cite{cui2026embridge}. Across sessions, electrode re-donning, changes in skin impedance, and movement variability alter the signal distribution~\cite{qiu2025placement}; across users, anatomical and neuromuscular differences introduce further shifts.

To handle these shifts, previous methods combined handcrafted features with linear discriminant analysis or support vector machines~\cite{hudgins1993strategy,englehart2003robust}. Although effective with limited data, their fixed features could not adapt to changes in the sensor interface or user physiology. Deep neural networks learn spatiotemporal features directly~\cite{hu2024vit, shabanpour2025moemba}, while domain adaptation and transfer learning further reduce source--target mismatch ~\cite{islam2024allconvnet,li2025endtoend}. Yet these methods focus on representations and decision boundaries rather than modeling the signal distribution, and they often require labeled target recordings for recalibration, which introduces extra user effort.

Generative models offer a complementary route by modeling the distribution of HD-sEMG signals~\cite{wang2026new}. DiffHGR uses diffusion for HD-sEMG augmentation and feeds multiscale denoising features into the decoding branch of an auxiliary autoencoder~\cite{diffhgr2026}. At inference, however, prediction uses only the autoencoder's encoder and a separately trained classifier; the diffusion is not used. This separation raises a question: \textit{Can the learned dynamics themselves make the class decision?} Flow matching (FM)~\cite{lipman2023flow, liu2022rectified} offers a potential route by learning a velocity field along predefined probability paths without trajectory simulation during training. However, its use in EMG remains limited to generation. EMGFlow~\cite{jiang2026emgflow} conditions noise-to-signal transport on a known gesture label, treating the label as an input rather than a prediction target and thus providing no direct classification rule. We therefore propose \textbf{MyoFlow}, to the best of our knowledge, the first FM framework for HD-sEMG classification across sessions and subjects. It recasts classification as subject- and session-conditioned rectified-flow transport from encoded windows to a shared bank of learnable gesture anchors. These anchors serve as both transport endpoints and decision prototypes, allowing classification through nearest-anchor matching without a separate classifier head. Meanwhile, the shared geometry enables zero-shot recognition across domains without target-domain labels. Our main contributions are:

\begin{figure*}[!t]
  \begin{center}
    \centerline{\includegraphics[width=0.9\linewidth]{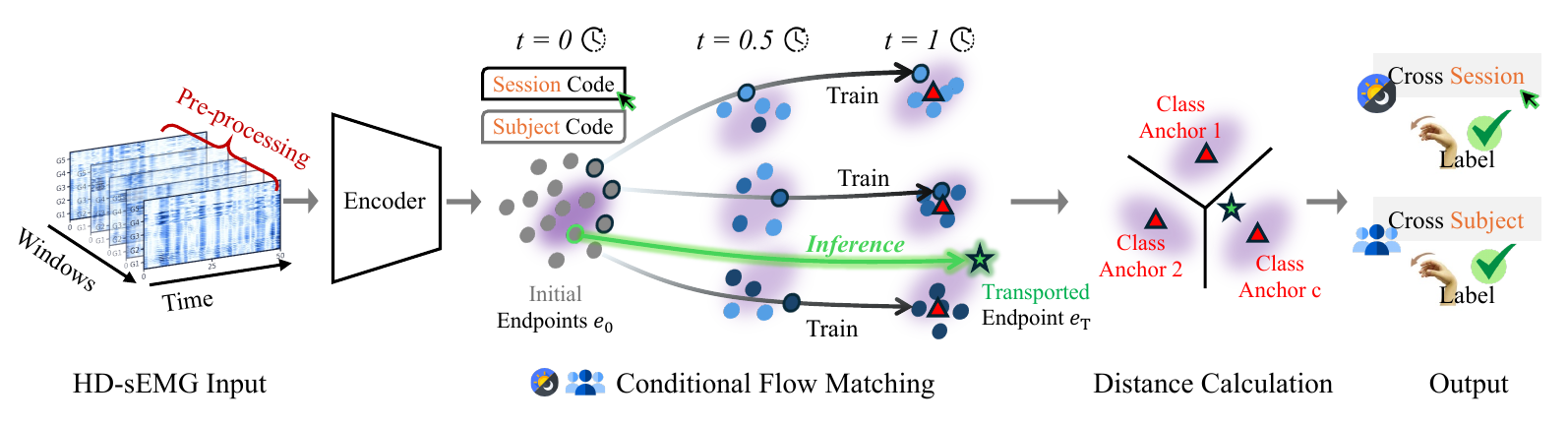}}
    \caption{
      The framework of the proposed MyoFlow for HD-sEMG gesture recognition across sessions and subjects.}
    \label{fig:framework}
  \end{center}
\end{figure*}

\ding{182} We propose \textbf{MyoFlow} for HD-sEMG gesture recognition, formulating cross-session and cross-subject recognition as rectified transport in the representation space.

\ding{183} We design an anchor-tied flow in which gesture anchors act as both transport targets and class prototypes, coupling representation learning to a nearest-anchor decision and making the learned class geometry identifiable.

\ding{184} \textbf{MyoFlow} improves recognition on multiple datasets across sessions and subjects in zero- and few-shot settings.

\section{Method}
\label{sec:method}

\subsection{Problem Formulation}
\label{sec:problem-overview}
Let \(x\in\mathbb{R}^{M\times L}\) denote an HD-sEMG window with \(M\) electrode channels and \(L\) temporal samples, labeled \(y\in\{1,\ldots,C\}\) and tagged with the subject \(p\) and session \(d\) of its recording. Training uses only the labeled source set \(\mathcal{D}_{\mathrm{src}}\). Cross-session recognition tests a source subject in a new session, and cross-subject recognition tests a subject absent from \(\mathcal{D}_{\mathrm{src}}\). Evaluation is indexed by the number \(K\) of labeled target repetitions released for calibration: \(K=0\) denotes zero-shot; the \(K\) class repetitions form the support set and the remainder the query set for $K > 0$. The split is made at the repetition level, since windows from one repetition are strongly correlated without using query labels. The proposed MyoFlow realizes recognition as an initial-value problem in a latent space of endpoint dimension \(D=256\):
\begin{equation}
\begin{aligned}
&e(0)=e_0=E_\theta(x),\quad e(1)=e_T,\\
&\frac{\mathrm{d}e(t)}{\mathrm{d}t}=v_\phi\!\left(e(t),t,c_{p,d}\right),
\end{aligned}
\label{eq:myoflow-overview}
\end{equation}
where \(E_\theta\) denotes the initial endpoint encoder, and \(v_\phi\) a velocity field conditioned on domain context \(c_{p,d}\). Following flow matching, \(t\in[0,1]\) is the interpolation coordinate of a probability path: \(t=0\) is the encoded window and \(t=1\) the target distribution. The terminal endpoint \(e_T=e(1)\) is thus where the class decision is read off. 

Fig.~\ref{fig:framework} provides an overview of our method. The anchor bank \(A=[a_1,\ldots,a_C]^\top\in\mathbb{R}^{C\times D}\) holds one learnable anchor per gesture, shared across subjects and sessions, and supplies the targets of the path. Since one anchor bank supplies both the flow's destinations and the classifier's prototypes, representation learning and the decision rule are optimized on a single geometry. The encoder \(E_\theta\) keeps time as the convolutional axis: a \(1\times1\) stem mixes the \(M\) electrodes into feature channels, a stack of dilated depthwise temporal blocks widens the receptive field without striding, and global average pooling with a two-layer projection gives \(e_0\in\mathbb{R}^{D}\).

\subsection{Conditional Rectified Flow}
\label{sec:rectified-flow}
For each labeled source window, MyoFlow pairs \(e_0\) with the anchor \(a_y\) of its label, samples \(t\sim\mathcal{U}(0,1)\), and defines the interpolated state and its constant target velocity as
\begin{equation}
e_t=(1-t)e_0+t\,\mathbf{sg}[a_y],
\qquad
u_t=\mathbf{sg}[a_y]-e_0,
\label{eq:rectified-path}
\end{equation}
with \(\mathbf{sg}[\cdot]\) the stop-gradient. The field is fitted by
\begin{equation}
\mathcal{L}_{\mathrm{FM}}=\mathbb{E}_{(x,y,p,d)\sim\mathcal{D}_{\mathrm{src}},\,t}\left\|v_\phi(e_t,t,c_{p,d})-u_t\right\|_2^2.
\label{eq:flow-matching-loss}
\end{equation}

As depicted in Fig.~\ref{fig:framework}, the label selects the target anchor but never enters \(v_\phi\), so the field stays usable when the label is unknown; its squared-error optimum is the conditional mean \(\mathbb{E}[u_t|e_t,t,c_{p,d}]\), averaging the target velocities of paths meeting at the same state. Since \(a_y\) is detached, the field learns to reach the anchors rather than drag them toward the data, while gradients through \(e_0\) shape the encoder and projector.

The domain context \(c_{p,d}\) is generated by a small Multi-Layer Perceptron (MLP) from concatenated subject and session embeddings, enabling domain-specific transport under shared anchor geometry. At test time, an unseen session reuses the source-session embedding, while an unseen subject uses the mean training-subject embedding; neither requires target domain data. Zero-initializing the output layer of \(v_\phi\) makes the initial transport map the identity. The FM loss requires no trajectory simulation, and the ordinary differential equation is solved only to obtain \(e_T\) for the anchor-tied classifier in Sec.~\ref{sec:anchor-classification}, using Euler integration with midpoint time sampling at four steps in training and eight in evaluation.

\subsection{Anchor-Tied Classification}
\label{sec:anchor-classification}
The anchor bank is initialized as orthonormal rows of a QR factorization, rescaled to a fixed radius \(r=\sqrt{D}\), so that all \(C(C-1)/2\) class pairs start equidistant. MyoFlow classifies the terminal endpoint \(e_T\) by its distance to each anchor:
\begin{equation}
\ell_c(e_T)=-\|e_T-a_c\|_2^2/\tau,\qquad \hat{y}=\arg\max\nolimits_{c}\ell_c(e_T),
\label{eq:anchor-readout}
\end{equation}
where \(\ell_c\) is the logit of class \(c\in\{1,\ldots,C\}\) and \(\tau=D/s_{\mathrm{anchor}}\) sets the logit scale, with \(s_{\mathrm{anchor}}=10\). Expanding the squared distance yields a tied linear read-out with class weight \(2a_c/\tau\) and bias \(-\|a_c\|_2^2/\tau\). Prediction is therefore equivalent to nearest-anchor classification. These weights are the flow's own destinations; therefore, an invertible reparameterization of the endpoint space can no longer be absorbed into them. Only a global isometry leaves the objective unchanged, and pairwise anchor distances are invariant under it. The learned class geometry is therefore explicit rather than an artifact of the read-out.

\subsection{Source-Only Learning}
\label{sec:source-learning}

No target-session window, statistic or label enters training. The source objective pairs flow matching with anchor supervision and auxiliary regularization:
\begin{equation}
\mathcal{L}_{\mathrm{train}}=\lambda_{\mathrm{FM}}\mathcal{L}_{\mathrm{FM}}+\mathcal{L}_{\mathrm{anchor}}+\sum_{r\in\mathcal{R}}\lambda_r\mathcal{L}_r,
\label{eq:training-objective}
\end{equation}
where \(\mathcal{L}_{\mathrm{anchor}}=\lambda_{\mathrm{CE}}\,\mathrm{CE}(\ell(e_T),y)+\lambda_{\mathrm{pull}}\|e_T-a_y\|_2^2/D\), and \(\mathcal{R}=\{\mathrm{geo},\mathrm{rad},\mathrm{con}\}\) collects regularizers on class geometry, anchor norm, and clean--augmented consistency. The two leading terms divide the supervision: \(\mathcal{L}_{\mathrm{FM}}\) constrains the path, \(\mathcal{L}_{\mathrm{anchor}}\) scores the arrival, and the latter is the only route by which the anchor bank is updated.

Optimization runs in two source-only stages: shared pretraining over source sessions pooled across subjects, then, for cross-session evaluation, adaptation on the evaluated subject's own source session. Leave-one-subject-out evaluation omits the second stage and excludes the held-out subject from pretraining, so that subject contributes no gradient at any point.

\subsection{Few-Shot Calibration}
\label{sec:fewshot-calibration}
For \(K>0\), calibration acts on \(e_0\) rather than \(e_T\), since transport concentrates \(e_T\) onto the anchors and leaves little within-class variation to correct. From the labeled support alone we recompute BatchNorm statistics and initialize a linear calibration head \(g_\omega\) from the class-wise support means \(\mu_c\), with \(w_c=\mu_c\) and \(b_c=-\|\mu_c\|_2^2/2\), which is exactly nearest-prototype classification. The encoder, projector, and \(g_\omega\) are then optimized on augmented support windows while the flow and anchor bank stay frozen; query repetitions are never seen. Let \(\mathcal{J}(z)=\mathrm{CE}(g_\omega(z),y)\), \(\varphi(x)=e_0\) and \(\varphi_j(x)\) denote the pooled endpoint and the projection of the pre-pooling feature map at position \(j\). The head \(g_\omega\) is induced and supervised at both resolutions:
\begin{equation}
\mathcal{L}_{\mathrm{cal}}=\mathbb{E}_{(x,y)}\Big[\mathcal{J}(\varphi(x))+\frac{\lambda_{\mathrm{frame}}}{L}\sum\nolimits_{j=1}^{L}\mathcal{J}(\varphi_j(x))\Big].
\label{eq:calibration-objective}
\end{equation}
The frame term draws \(L\) supervised signals from each window. And since the projector is non-linear, \(g_\omega\) receives gradients from features that the pooled term never produces.

\begin{table}[t]
\centering
\caption{Cross-\textbf{session} accuracy on Hyser (\%).}
\label{tab:cross-session}
\small
\setlength{\tabcolsep}{3.0pt}
\resizebox{\columnwidth}{!}{%
\begin{tabular}{@{}lccc|>{\columncolor{creamavg}}c@{}}
\toprule
\textbf{Method} & \textbf{0-rep (ZS)} & \textbf{1-rep (FS)} & \textbf{2-rep (FS)} & \textbf{Avg.} \\
\midrule
ViT-MDHGR~\cite{hu2024vit}(2024) & 72.93 & 83.47 & 86.13 & 80.84 \\
MoEMba~\cite{shabanpour2025moemba}(2025) & 58.45 & 72.55 & 76.87 & 69.29 \\
DiffHGR~\cite{diffhgr2026}(2026) & \textcolor{blue}{\underline{76.32}} & \textcolor{blue}{\underline{85.70}} & \textcolor{blue}{\underline{87.34}} & \textcolor{blue}{\underline{83.12}} \\
\rowcolor{gray!10}
\textbf{MyoFlow (ours)} & \textcolor{red}{\textbf{79.19}} & \textcolor{red}{\textbf{89.80}} & \textcolor{red}{\textbf{93.08}} & \cellcolor{creamavgdark}\textcolor{red}{\textbf{87.36}} \\
\bottomrule
\end{tabular}}
\end{table}

\begin{table}[t]
\centering
\caption{Cross-\textbf{subject} accuracy on Hyser (\%).}
\label{tab:cross-subject}
\small
\setlength{\tabcolsep}{3.0pt}
\resizebox{\columnwidth}{!}{%
\begin{tabular}{@{}lccc|>{\columncolor{creamavg}}c@{}}
\toprule
\textbf{Method} & \textbf{0-rep (ZS)} & \textbf{1-rep (FS)} & \textbf{2-rep (FS)} & \textbf{Avg.} \\
\midrule
ViT-MDHGR~\cite{hu2024vit}(2024) & 60.90 & 76.71 & \textcolor{blue}{\underline{80.74}} & 72.78 \\
MoEMba~\cite{shabanpour2025moemba}(2025) & 46.76 & 66.35 & 72.36 & 61.82 \\
DiffHGR~\cite{diffhgr2026}(2026) & \textcolor{blue}{\underline{63.37}} & \textcolor{blue}{\underline{78.01}} & 80.30 & \textcolor{blue}{\underline{73.89}} \\
\rowcolor{gray!10}
\textbf{MyoFlow (ours)} & \textcolor{red}{\textbf{64.12}} & \textcolor{red}{\textbf{86.55}} & \textcolor{red}{\textbf{90.11}} & \cellcolor{creamavgdark}\textcolor{red}{\textbf{80.26}} \\
\bottomrule
\end{tabular}}
\end{table}

\begin{figure}[t]
  \begin{center}
    \centerline{\includegraphics[width=0.9\linewidth]{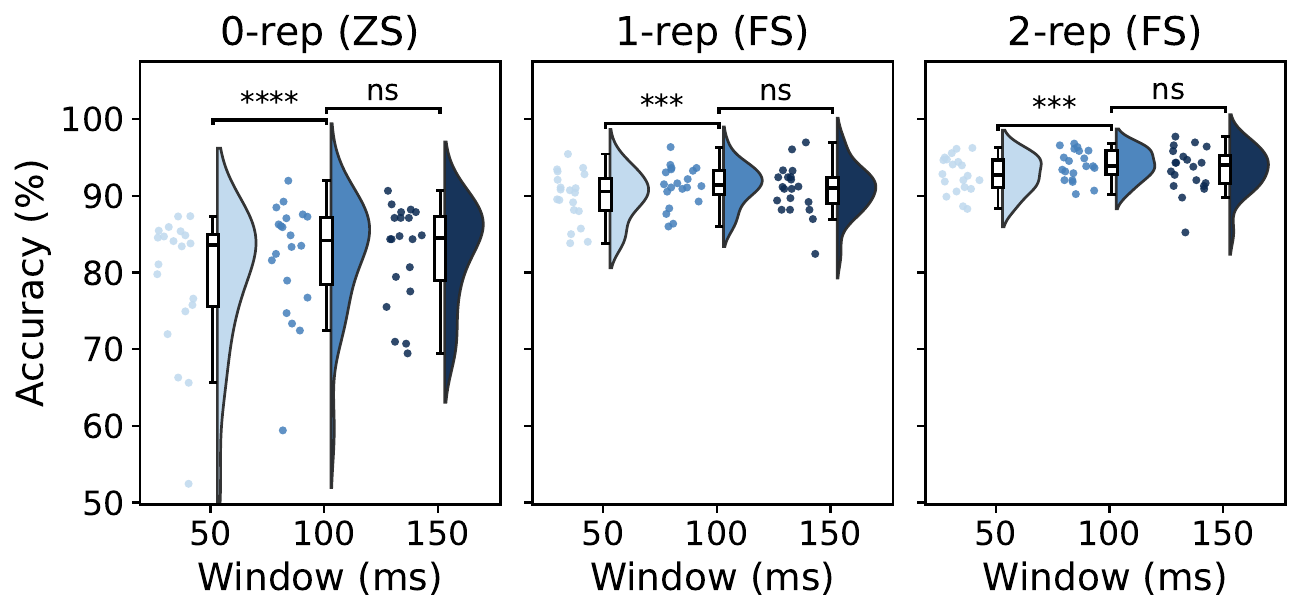}}
    \caption{
      Temporal-window sensitivity for cross-session recognition on Hyser, with pairwise significance tests.}
    \label{fig:hyser_window}
  \end{center}
\end{figure}

\section{Experiment Setup}
\label{sec:experiment}

% ---------------- CEMHSEY cross-day table (full width) ----------------
\begin{table*}[!t]
\centering
\caption{Cross-\textbf{session} accuracy on CEMHSEY (\%) for day gaps of 1--10 days. (\textcolor{red}{\textbf{Best}} performance; \textcolor{blue}{\underline{Second best}} performance)}
\label{tab:cemhsey_crossday}
\small
\setlength{\tabcolsep}{3pt}
\renewcommand{\arraystretch}{1.05}
\resizebox{\textwidth}{!}{%
\begin{tabular}{@{}l *{3}{ccccc|>{\columncolor{creamavg}}c}@{}}
\toprule
\multirow{2}{*}{\textbf{Method}}
 & \multicolumn{6}{c}{\textbf{0-rep (ZS)}}
 & \multicolumn{6}{c}{\textbf{1-rep (FS)}}
 & \multicolumn{6}{c}{\textbf{2-rep (FS)}} \\
\cmidrule(lr){2-7} \cmidrule(lr){8-13} \cmidrule(lr){14-19}
 & Day 2 & Day 4 & Day 6 & Day 8 & Day 11 & \textbf{Avg.}
 & Day 2 & Day 4 & Day 6 & Day 8 & Day 11 & \textbf{Avg.}
 & Day 2 & Day 4 & Day 6 & Day 8 & Day 11 & \textbf{Avg.} \\
\midrule
ViT-MDHGR~\cite{hu2024vit} (2024)
 & 90.57 & \textcolor{blue}{\underline{86.63}} & 85.97 & 84.05 & 83.00 & 86.04
 & 96.38 & 93.62 & 94.75 & 96.62 & 94.87 & 95.25
 & 97.38 & 95.28 & 96.26 & 96.57 & 96.63 & 96.42 \\
MoEMba~\cite{shabanpour2025moemba} (2025)
 & 78.88 & 72.60 & 70.69 & 68.87 & 68.22 & 71.85
 & 89.34 & 87.63 & 88.48 & 89.94 & 87.92 & 88.66
 & 91.31 & 91.21 & 92.42 & 93.47 & 92.94 & 92.27 \\
DiffHGR~\cite{diffhgr2026} (2026)
 & \textcolor{blue}{\underline{92.69}} & 85.56 & \textcolor{blue}{\underline{87.90}} & \textcolor{blue}{\underline{84.80}} & \textcolor{blue}{\underline{86.67}} & \textcolor{blue}{\underline{87.52}}
 & \textcolor{blue}{\underline{96.46}} & \textcolor{blue}{\underline{94.44}} & \textcolor{red}{\textbf{96.81}} & \textcolor{blue}{\underline{97.73}} & \textcolor{blue}{\underline{97.14}} & \textcolor{blue}{\underline{96.52}}
 & \textcolor{blue}{\underline{97.40}} & \textcolor{blue}{\underline{96.42}} & \textcolor{red}{\textbf{97.69}} & \textcolor{blue}{\underline{97.39}} & \textcolor{blue}{\underline{97.89}} & \textcolor{blue}{\underline{97.36}} \\
\rowcolor{gray!10}
\textbf{MyoFlow (ours)}
 & \textcolor{red}{\textbf{96.06}} & \textcolor{red}{\textbf{90.84}} & \textcolor{red}{\textbf{91.87}} & \textcolor{red}{\textbf{89.66}} & \textcolor{red}{\textbf{90.11}} & \cellcolor{creamavgdark}\textcolor{red}{\textbf{91.71}}
 & \textcolor{red}{\textbf{97.41}} & \textcolor{red}{\textbf{94.92}} & \textcolor{blue}{\underline{96.77}} & \textcolor{red}{\textbf{98.22}} & \textcolor{red}{\textbf{97.42}} & \cellcolor{creamavgdark}\textcolor{red}{\textbf{96.95}}
 & \textcolor{red}{\textbf{98.07}} & \textcolor{red}{\textbf{97.14}} & \textcolor{blue}{\underline{97.37}} & \textcolor{red}{\textbf{98.01}} & \textcolor{red}{\textbf{98.55}} & \cellcolor{creamavgdark}\textcolor{red}{\textbf{97.83}} \\
\bottomrule
\end{tabular}}
\end{table*}

We evaluate MyoFlow on the Hyser PR Dynamic~\cite{jiang2021open} (20 subjects, two sessions separated by 3–25 days, 11 gestures, and 256 channels) and the CEMHSEY~\cite{yang2025consecutive} (6 subjects, 11 consecutive days, 11 gestures, and 320 channels). Both datasets are sampled at 2048~Hz and segmented into non-overlapping 50~ms windows. Cross-session evaluation uses Hyser Session~1 as the source and Session~2 as the target, while CEMHSEY uses Day~1 as the source and Days~2, 4, 6, 8, and 11 as targets. Cross-subject evaluation follows 20-fold leave-one-subject-out validation on Hyser Session~1, with no adaptation to the held-out subject. For \(K\in\{0,1,2\}\), \(K\) complete target repetitions per class form the support set and the remainder form the query set. All experiments are conducted on a single NVIDIA RTX~5090 GPU.

\begin{figure}[!t]
  \begin{center}
    \centerline{\includegraphics[width=\linewidth]{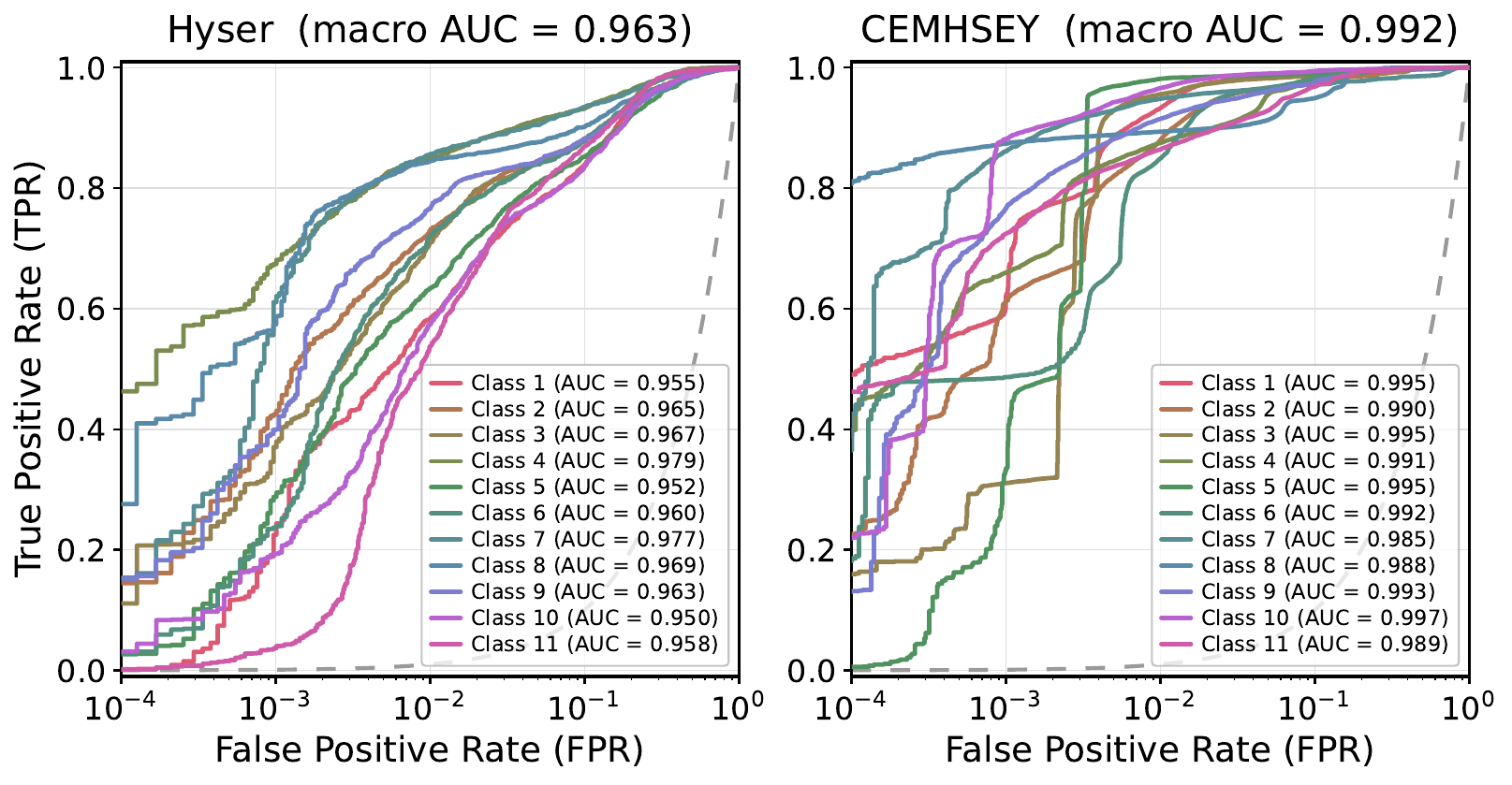}}
    \caption{
      Zero-shot Receiver Operating Characteristic (ROC) curves of MyoFlow on both datasets. (The FPR is log-scaled) }
    \label{fig:roc}
  \end{center}
\end{figure}

\section{Results and Analysis}

\textbf{Overall comparison.}
As shown in Tables~\ref{tab:cross-session} and~\ref{tab:cross-subject}, MyoFlow outperforms every listed baseline across all Hyser settings. Relative to DiffHGR, the strongest baseline by average accuracy, it improves cross-session and cross-subject performance by 4.24 and 6.37 percentage points, respectively. For unseen subjects, MyoFlow exceeds the best 1- and 2-repetition baselines, DiffHGR and ViT-MDHGR, by 8.54 and 9.37 points. Table~\ref{tab:cemhsey_crossday} further shows that MyoFlow achieves the highest zero-shot accuracy on every CEMHSEY target day, increasing the average from 87.52\% to 91.71\%. The benefit of 2-repetition calibration over its zero-shot result grows from 2.01 points on Day~2 to 8.44 points on Day~11. These findings reflect the complementary roles of the two stages: anchor-tied rectified transport establishes a shared decision geometry for zero-shot recognition, while initial-endpoint calibration uses target support to correct residual distribution shift.

\textbf{Temporal sensitivity.}
Fig.~\ref{fig:hyser_window} shows that increasing the window length from 50 to 100\,ms improves accuracy across all settings, as confirmed by paired Wilcoxon signed-rank tests, whereas extending it to 150\,ms yields no significant further gain. These results reveal an accuracy--latency trade-off: compared with 100\,ms, the 50\,ms setting halves acquisition time while retaining strong recognition accuracy, supporting deployment in latency-sensitive settings.

\begin{figure}[!t]
  \begin{center}
    \centerline{\includegraphics[width=\linewidth]{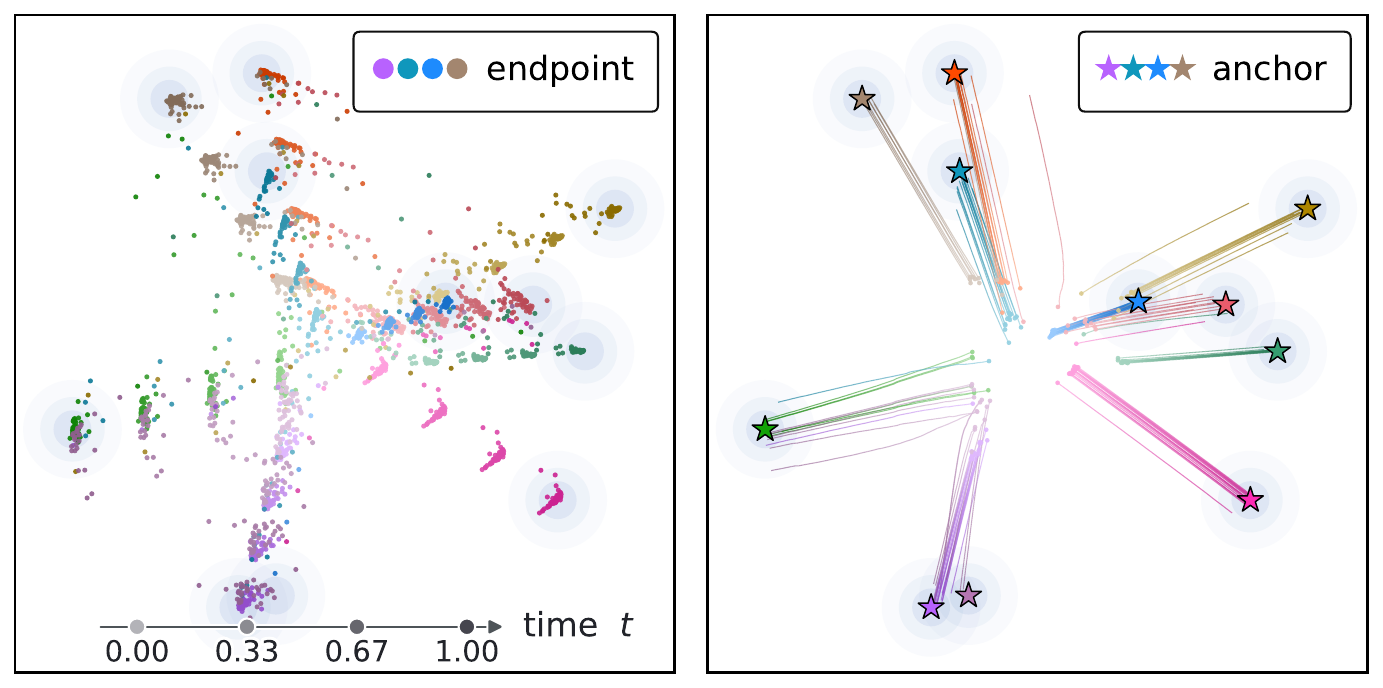}}
    \caption{
      Visualization of the rectified-flow transport that carries each initial endpoint to its class anchor.}
    \label{fig:flow}
  \end{center}
\end{figure}

\textbf{Classifier discrimination.}
We construct class-wise one-vs-rest ROC curves by pooling target windows within each dataset and sweeping thresholds over the per-window, zero-shot, anchor-derived probabilities. Fig.~\ref{fig:roc} reports macro-AUCs of 0.963 on Hyser and 0.992 on CEMHSEY, with every class-wise AUC exceeding 0.950. These results show that the anchor-tied readout maintains reliable class-wise discrimination under both cross-session and cross-day distribution shifts, complementing the multiclass accuracy results.

\textbf{Interpretability of learned transport.}
Fig.~\ref{fig:flow} visualizes the endpoint trajectories projected onto the two-dimensional principal subspace of the anchor bank. As flow progresses, the endpoints move toward and concentrate around their class anchors, revealing how rectified transport organizes the latent space into a decision-aligned geometry. Because prediction uses the same anchors, each trajectory directly connects representation dynamics to the final decision, providing a geometric, sample-level interpretation of recognition.

\section{Conclusion}
We introduced MyoFlow, the first discriminative flow matching framework for cross-session and cross-subject HD-sEMG gesture recognition. Its anchor-tied design embeds transport in the decision rule: a domain-conditioned rectified flow moves encoded windows toward gesture anchors serving as transport endpoints and prototypes, enabling nearest-anchor prediction without an independent head. This geometry supports zero-shot transfer, while few-shot calibration adapts the variation-preserving initial endpoint to residual distribution shifts. Experiments on Hyser and CEMHSEY show strong recognition under session and subject shifts, with the largest calibration gains at longer day gaps. Together, these results show that learned transport can define the class decision rather than remain auxiliary to a separate classifier.

\section*{Acknowledgement}
This work was supported by the Waseda University Grant for Special Research Projects (Project No. BARH02612801), the Waseda Research Institute for Science and Engineering, and JST BOOST under Grant Number JPMJBS2429.

% \section{REFERENCES}
% \label{sec:refs}

\bibliographystyle{IEEEbib}
\bibliography{strings,refs}

\end{document}